\documentclass[sigconf,nonacm]{acmart}

\usepackage{algorithm}
\usepackage{algorithmic}
\usepackage{graphicx}
\usepackage{subfigure}
\usepackage{color}

\AtBeginDocument{%
  \providecommand\BibTeX{{%
    \normalfont B\kern-0.5em{\scshape i\kern-0.25em b}\kern-0.8em\TeX}}}

\begin{document}

\title{End-To-End Graph-based Deep Semi-Supervised Learning}



\author{Zihao Wang, Enmei Tu and Meng Zhou}
\affiliation{%
	\institution{Shanghai Jiao Tong University}
	\city{Shanghai}
	\state{China}}


\begin{abstract}
The quality of a graph is determined jointly by three key  factors of the graph: nodes, edges and similarity measure (or edge weights), and is very crucial to the success of graph-based semi-supervised learning (SSL) approaches. Recently, dynamic graph, which means part/all its factors are dynamically updated during the training process, has demonstrated to be promising for graph-based semi-supervised learning. However, existing approaches only update part of the three factors and keep the rest manually specified during learning stage.  In this paper, we propose a novel graph-based semi-supervised learning approach to optimize all three factors simultaneously in an end-to-end learning fashion. To this end, we concatenate two neural networks (feature network and similarity network) together to learn the categorical label and semantic similarity, respectively, and train the networks to minimize a unified SSL objective function. We also introduce an extended graph Laplacian regularization term to increase training efficiency. Extensive experiments on several benchmark datasets demonstrate the effectiveness of our approach. 
\end{abstract}

%



\maketitle

\section{Introduction}

Deep neural networks trained with a large number of labeled samples have attained tremendous successes in many areas such as computer vision, natural language processing and so on~\cite{he2016deep,huang2017densely,devlin2019bert}. However, labeling numerous data manually is expensive for many tasks (e.g. medical image segmentation) because the labeling work is often resource- and/or time-consuming. Semi-supervised learning (SSL), which leverages a small set of high quality labeled data in conjunction with a large number of easily available unlabeled data, is a primary solution to deal with this problem.  Comprehensive introductions and reviews of existing SSL approaches could be found in ~\cite{zhu2005semi,van2019survey}. 

In recent years, deep semi-supervised learning (DSSL) become an active research topic and  a surge of novel approaches appears in the literature. Broadly speaking, these DSSL approaches could be divided into two groups:

\begin{itemize}
	\item Pairwise similarity independent. This type of SSL approach uses sample-wise regularization into the model to reduce network overfitting on the small labeled dataset. Typical algorithms include:  forcing the perturbed version of the data or model to be  close to the clean version  \cite{bachman2014learning,sajjadi2016regularization,laine2017temporal,tarvainen2017mean,miyato2018virtual},  adopting generative model (mainly generative adversarial network (GAN) or autoencoder) to learn data distribution information ~\cite{odena2016semi,dai2017good,kumar2017semi,springenberg2016unsupervised,rasmus2015semi,kingma2014semi,xu2017variational} ,  utilizing pseudo label to expand the training set ~\cite{lee2013pseudo,yan2016robust,berthelot2019mixmatch,verma2019interpolation}.
	
	\item Pairwise similarity dependent. This type of SSL approach makes use of pair-wise relationships between all samples to reduce model overfitting. Typical algorithms include: combining with traditional graph-based SSL ~\cite{kamnitsas2018semi,taherkhani2019matrix,iscen2019label}, exploiting geometrical properties of data manifolds ~\cite{rifai2011manifold,weston2012deep,qi2018global}, embedding association learning ~\cite{haeusser2017learning}, deep metric embedding ~\cite{hoffer2016semi}, etc.
	
\end{itemize}

We focus on the pairwise similarity based SSL, in particular, graph-based approaches. In these approaches, a graph $G(V, E, W)$ (with vertex set $V$, edge set $E$ and edge weight/similarity measure $W$) is constructed and learning (i.e. label propagation or random walk) is performed on the graph. The quality of the graph (in terms of the connectivity of different components corresponding to different categorical classes in a classification problem) is controlled by its ingredients $V,E,W$ and is a dominant factor of a graph-based SSL to achieve good performance.  

\textit{Static-graph-based} approaches construct a  graph with constant $V, E, W$  and the graph remain fixed during model training. For these approaches, $V$ could be the raw sample set or transformed feature vectors of the raw samples, including training and testing samples. $E$ is usually generated by $k$-nearest-neighbor method or $\epsilon$ distance method\footnote{If the distance between two samples is less than $\epsilon$, there will be an edge between them.}. If $k$ equals sample set size or $\epsilon$ equals sample set diameter, the graph will be a complete graph. The edge weight $W$ is usually obtained by a predefined similarity measure function, $W_{ij}=f(v_i, v_j)$ (such as Gaussian kernel function or dot product), to reflect the affinity/closeness of a pair of vertices. Most traditional graph-based SSL (e.g. \cite{zhu2003semi,zhou2004learning,belkin2006manifold}) and early DSSL (e.g. \cite{weston2012deep}) utilize static graph.

\textit{Dynamic-graph-based} approaches, in contrast, construct a graph whose vertices and/or edges are continually updated during model training. Compared to static graph, dynamic graph could absorb the newest categorical information extracted by the classifier (e.g. a neural network) and thus adjust its structure to adapt the learning process immediately. Recent progress in SSL has demonstrated that dynamic graph is more advantageous and preferable for DSSL \cite{kamnitsas2018semi,luo2018smooth,iscen2019label}, because for complex learning task, such as natural image classification, any predefined static graph is either futile or arduous.

Existing dynamic-graph-based DSSL approaches usually adopt a specific weight/similarity function to construct the graph. In \cite{iscen2019label,kamnitsas2018semi} the authors use a dot product function to measure the similarity of hidden layer features. Luo \text{et al} assign 0 or 1 to the graph edges according to the pseudo labels of the corresponding network output.  The similarity measure methods in these approaches are not \textbf{\textit{trainable}} and thus may limit the adaptive capability of the dynamic graph, because edge similarity also has a direct influence upon graph quality, hence the performance of the SSL.

\begin{figure}[htbp!]
	\centering
	\includegraphics[width=0.48\textwidth]{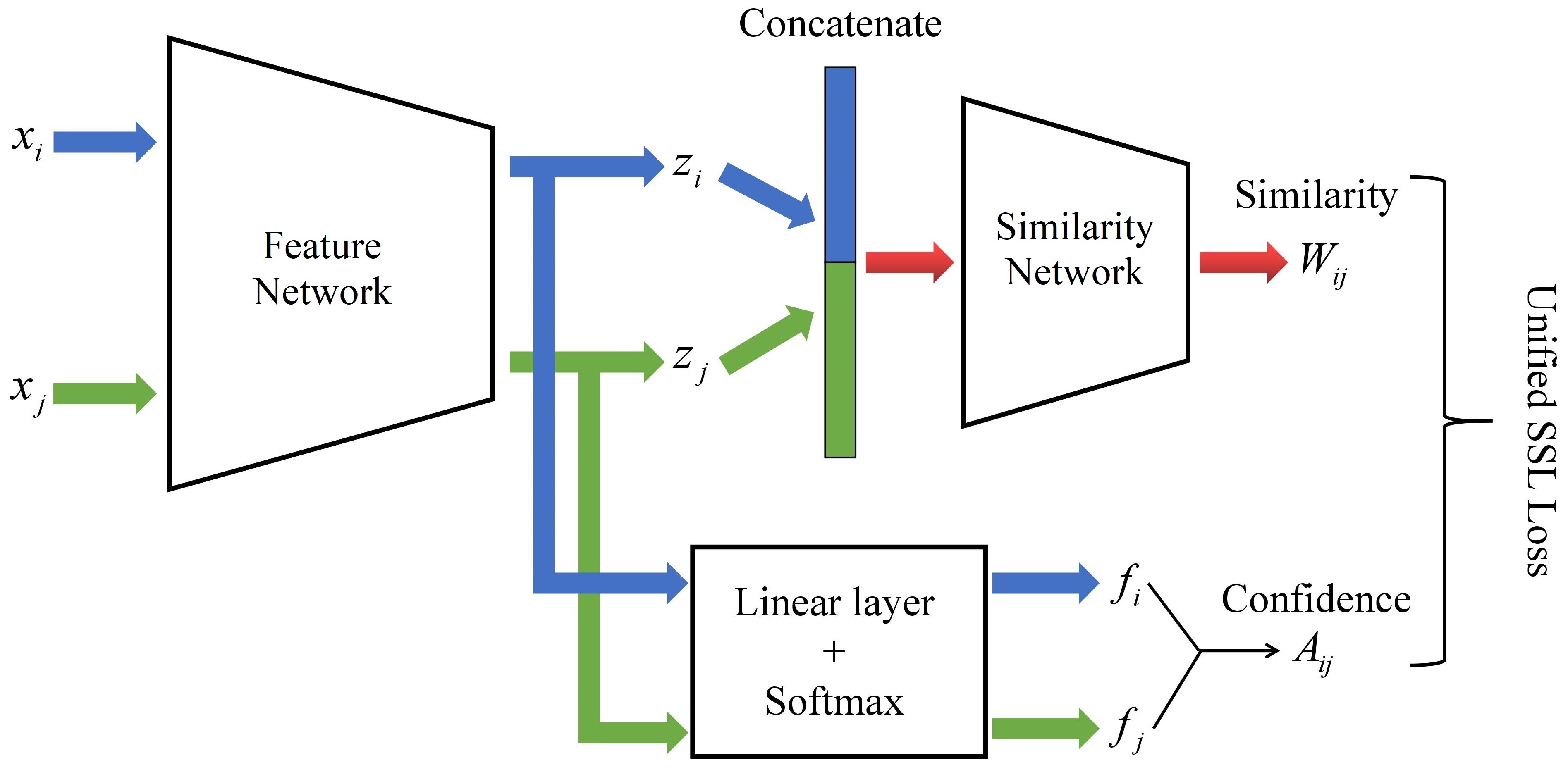}
	\caption{The architecture of our model. Feature network maps input samples to a latent space and similarity network learns the semantic similarity function in the latent space. The two networks are optimized jointly to minimize  a semi-supervised objective function.}
	\label{fig1}
\end{figure}

In this paper, we blend dynamic graph construction with a learnable similarity measure into a network and train a system consisting of two networks in an end-to-end fashion. In particular, our model contains two networks, feature network and similarity network, as shown in Figure~\ref{fig1}. The feature network maps input samples into a hidden space and, meanwhile, learns a classifier under the guidance sample labels. The similarity network learns a  similarity function in the hidden space also under the guidance of sample labels and its output value  is used to construct a dynamic graph to train the classifier.  Two networks are optimized together to minimize a novel unified SSL objective function.  Both networks' learning targets are the sample labels, so our model is an end-to-end learning approach.

\section{Related Works}

\subsection{Graph-based SSL}

A static graph means that once the matrix $A$ is computed (by a predefined similarity function or by locally adaptive methods). Once constructed, the graph remains constant during learning process. Traditional graph-based SSL algorithms usually contain two steps: a graph is constructed from both labeled and unlabeled samples; categorical information is propagated from labeled samples to unlabeled ones on the graph. Representative algorithms include graphcut \cite{blum2001learning}, label propagation (LP) \cite{xiaojin2002learning}, harmonic function (HF) \cite{zhu2003semi}, local and global consistency (LGC) \cite{zhou2004learning} and many others. Since graph and manifold has a close mathematical relationship, there are also graph-based algorithms exploiting differential geometry theory, such as manifold regularization \cite{belkin2006manifold}, manifold tangent \cite{rifai2011manifold}, Hessian energy \cite{kim2009semi}, local coordinate \cite{yu2009nonlinear}. Because of the importance of graph quality in graph-based SSL, researchers have also developed various techniques to optimize graph weight or transform sample features to obtain a better graph  \cite{leistner2008semi,cheng2009sparsity,jebara2009graph,li2013low}.

 However, SSL classification results could vary largely for different similarity matrices \cite{chapelle2009semi,zhu2009introduction}. These static-graph-based approaches may achieve great success on conventional classification tasks but rarely on complex tasks, such as natural image classification, because for these tasks it is almost impossible to generate a static graph which could faithfully capture all classification related information. 
 



In recent years, combining traditional graph-based SSL with deep neural networks to reduce training data demand has been an  active research topic. Perhaps, the earliest attempt is made by Weston \textit{et al.} \cite{weston2008deep} by including a graph Laplacian regularization term into the objective function. In \cite{kamnitsas2018semi}, a  graph is constructed in hidden feature space and the traditional graph-based SSL algorithm label propagation \cite{xiaojin2002learning} is adopted to compute the  CCLP (Compact
Clustering via Label Propagation) regularizer. In \cite{iscen2019label}, a  graph is constructed on the hidden features of a  training batch and the LGC \cite{zhou2004learning} is used to obtain pseudo labels, which are treated as ground truth to train the network in the next round. Taherkhani \textit{et at.} use matrix completion to predict labeling matrix and construct a graph in hidden feature space to minimize the triplet loss of their network \cite{taherkhani2019matrix}.   In \cite{luo2018smooth}, Luo \textit{et al}.~\shortcite{luo2018smooth} utilize the marginal loss  \cite{hadsell2006dimensionality} to exert neighborhood smoothness on a 0-1 sparse dynamic graph for each mini-batch. Different from \cite{weston2008deep}, the graph in these algorithms dynamically evolves in the training stage. Due to the strong adaption property of the dynamic graph, these semi-supervised learning approaches have shown appealing performance for complex classification tasks.

\subsection{Perturbation-based SSL}


These methods force two related copies of an individual sample, e.g. an image and its augmented version, to have consistent network outputs. The so-called consistency regularization term is defined in equation (\ref{equ1}).
\begin{equation}
\label{equ1}
L_{c}(x;\theta)=\sum_{i}l_{c}
(f_{\theta}(x_{i}),f_{\tilde{\theta}}(\tilde{x}_{i}))
\end{equation}%
where $ \tilde{x}_{i} $ is a transformation of sample $ x_{i} $ and parameter set $ \tilde{\theta} $ is either equal to $ \theta $ or any other transformation of it. $ f $ is the classification output distribution. Perturbation difference $ l_{c} $ is commonly measured by the squared Euclidean distance, i.e. $ \|f_{\theta}(x_{i}) - f_{\tilde{\theta}}(\tilde{x}_{i})\|^{2} $.

The perturbations in $ \Pi $ model~\cite{laine2017temporal} include data augmentation, input Gaussian noise and network's dropout, etc. Temporal Ensembling model in~\cite{laine2017temporal} forces the outputs of the current network to learn temporal average values during training. Mean teacher (MT)~\cite{tarvainen2017mean} averages network parameters to obtain an online and more stable target $ f_{\tilde{\theta}}(x) $. Besides, in virtual adversarial training (VAT)~\cite{miyato2018virtual}, adversarial perturbation which maximally changes the output class distribution acts as an effective perturbation in the consistency loss. We will adopt perturbation loss in equation (\ref{equ1}) into our model as a regularization term to reduce overfitting.


%

\section{Our Method}
For graph-based semi-supervised learning, feature extraction and similarity measure are mutually beneficial to each other, i.e. better feature yields better similarity measure and  vice versa.  As displayed in Figure \ref{fig1}, we propose a joint architecture to simultaneously optimize feature extraction  and similarity learning  to minimize a novel SSL objective function.  We also introduce an extended version of traditional graph Laplacian regularized term \cite{belkin2006manifold}  to prevent the trivial-solution problem (always output 0 regardless of input) for traditional graph Laplacian \cite{weston2012deep}. The overall objective function contains supervised and unsupervised loss parts for each component network.  First, let us introduce some mathematical notations.

Given a data set $ \mathcal{X} = \{x_{1}, x_{2}, ..., x_{l}, x_{l+1}, ..., x_{n}\}  $ with $ x_{i} \in \mathcal{R}^{d} $, SSL assumes the first $ l $ samples $ \mathcal{X}_{L} = \{x_{1}, x_{2}, ..., x_{l}\} $ are labeled according to $ \mathcal{Y}_{L} = \{y_{1}, y_{2}, ..., y_{l}\} $ with $ y_{i} \in C = \{1, 2, ..., c\} $ and the rest $ n-l $ samples $ \mathcal{X}_{U} = \{x_{l+1}, ..., x_{n}\} $ are unlabeled (usually $ l \ll n $). The binary label matrix of $\mathcal{Y}_L$ is denoted as $Y$, whose elements are $y_{ij}=1$ if and only if sample $x_i$ is from class $j$. The goal of SSL is to learn a classifier $ f: \mathcal{X} \rightarrow [0,1]^{c} $ parameterized by $ \theta $ using all samples in $ \mathcal{X} $ and the labels $ \mathcal{Y}_{L} $. It is usually solved by minimizing equation (\ref{equ3}).
\begin{equation}
\label{equ3}
\mathop {\min}\limits_\theta\sum_{i=1}^{l}L_{s}(f_{\theta}(x_{i}), y_{i}) + L_{u}(f_{\theta}(\mathcal{X}_{L}, \mathcal{X}_{U}))
\end{equation}%
where $ L_{s} $ is a supervised loss term, e.g. mean squared error (MST) or cross-entropy loss. $ f_{\theta}(x) $ is the parameterized classifier and $ L_{u} $ is usually a regularization term to exploit the unlabeled samples’ information. To encourage categorical information distributed smoothly, a graph Laplacian regularization term is adopted to penalize abrupt changes of the labeling function $f$ over nearby samples 
\begin{equation}
\label{equ2}
L_u=\sum_{i,j}A_{ij}\|f(x_{i})-f(x_{j})\|^{2} = f^T\Delta f
\end{equation}%
where $ A_{ij} $ denotes the pairwise similarity between samples $ x_{i} $ and $ x_{j} $ and $A=\{A_{ij}\}_{i,j=1}^{n}$ is the affinity/similarity matrix to encode nodes closeness on the graph. $\Delta=D-A$ is the graph Laplacian matrix\footnote{One could equally use the normalized version $\bar{\Delta}=I-D^{-1/2}AD^{-1/2}$. Here $I$ is the identity matrix of proposed size.} and $D$ is a diagonal matrix with $D_{ii}=\sum_{j}A_{ij}$.

\subsection{The Supervised Part}

\subsubsection{Learning Categorical Labels}

We use a deep convolutional neural network as the feature classifier  $ f $. It can be decomposed as $ f = h \circ g $, where $ g: \mathcal{X} \rightarrow \mathcal{R}^{d} $ is a feature extractor which maps the input samples to abstract features; and $ h: g(x)\rightarrow [0,1]^{c} $ is a linear classifier which is always connected by a softmax function to output the probability distribution for each class. We denote the feature extracted from sample $ x_i $ by $ z_i = g(x_i) $ here.

For categorical labels learning, we use the standard cross-entropy loss as the supervised loss term, as shown in equation (\ref{equ4}).
\begin{equation}
\label{equ4}
L_{sup\_f} = -\frac{1}{l}\sum_{i=1}^{l}\sum_{j=1}^{c}y_{ij}ln(f_{\theta}(x_i)_j)
\end{equation}%
Note that $ L_{sup\_f} $ only applies to labeled samples in $  \mathcal{X}_L $.

\subsubsection{Learning  Semantic Similarity}

Since $ z $ is a low-dimensional feature of the input $x$, the similarity $ W_{ij} $ between two samples $ x_i $ and $ x_j $ can be formulated as a function of latent variables $ z_i, z_j$, as shown in equation (\ref{equ5}).
\begin{equation}
\label{equ5}
W_{ij} = \Phi(z_i, z_j) = \Phi(g(x_i), g(x_j))
\end{equation}%
where $ \Phi(\cdot) $ is a multilayer fully connected neural network (similarity network). Equation (\ref{equ5}) also shows that pairwise similarity $ W_{ij} $ is a composite function of sample pair $ x_i, x_j $.

To construct an end-to-end semantic similarity learning model, we consider the task of learning  $ W_{ij} $ as a binary supervised classification problem. So, there are two units in the output of $ \Phi(\cdot) $, which represent similarity and dissimilarity respectively. Since the network's output after softmax is a probability distribution, if the similarity between two samples is $ W_{ij} $, then the dissimilarity between them is $ (1-W_{ij}) $. We also use the cross-entropy loss as a supervised loss term to learn the semantic similarity, as shown in equation (\ref{equ6}).
\begin{equation}
\label{equ6}
L_{sup\_W} = -\frac{1}{l^2}\sum_{ij}\sum_{k=1}^{2}W_{ijk}^{l}ln(\Phi(z_i,z_j)_k)
\end{equation}%
where $ W_{ij}^{l} \in [0,1]^2 $ is the similarity target vector, whose value is $ [1,0] $ if the sample pairs $x_i,x_j \in \mathcal{X}_L$ from the same class and $ [0,1] $ if from different classes. However, dissimilar pairs are far more than similar pairs and if we randomly select sample pairs from the labeled set $\mathcal{X}_L$, model can hardly learn from similar pairs. To tackle with this problem,  we generate virtual pairs including $ x_i $ and its augmented version $ Augment(x_i) $ as a similar pair in a mini-batch during training.  Notice that $x_i$ and $ Augment(x_i) $ in generated virtual pairs are similar but $ z_i, z_j $ are not equal because of the random data augmentation, including Gaussian noise and network's dropout, etc. We should mention that in our model, the task of learning similarity $ W $ here is parallel to the task of learning label $ y $ and has equal importance.

\subsection{The Unsupervised Part}
The supervised losses in equations (\ref{equ4}) and (\ref{equ6}) are defined on the labeled set $\mathcal{X}_L$. Now we introduce the unsupervised loss term in equation (\ref{equ3}).

\subsubsection{An Extended Graph Laplacian Regularization}

The categorical probability distribution given by $f$ reflects how likely the classifier regards a sample coming from each class. Therefore, we could define a classifier confidence over two samples as 
\begin{equation}
\label{equ7}
A_{ij} = \exp(-\beta\|f(x_i)-f(x_j)\|^2)
\end{equation}%
where $ A_{ij} $ is the classifier confidence that   samples $ x_i $ and $ x_j $ are from the same class. The confidence is also a similarity measure for samples $ x_i $ and $ x_j $, and, contrarily, $ (1-A_{ij}) $ represents the dissimilarity between the samples. By doing so, we construct a classifier confidence graph $G_c(\{f_i\}, {A_{ij}})$, whose nodes are categorical probability distribution of the samples and whose edge weights are the same-class confidence $A_{ij}$.


Trained by equation (\ref{equ6}), the semantic similarity  $ \Phi(z_i, z_j) $ given by similarity network  can be served as the ideal similarity value. We construct a semantic graph $G_s(z_i, \Phi)$ on hidden layer features with nodes $z_i, i=1...n$ and edge weights $\Phi(z_i, z_j)$. The purpose of regularizing unlabeled samples is to optimize the confidence graph $G_c$ towards the semantic graph $G_s$. We encourage the two graphs matching each other by minimizing cross entropy between confidence similarity $A_{ij}$ and 
semantic similarity  $ \Phi(z_i, z_j) $ 
\begin{align}
\label{equ8}
L_{unsup} = -\sum_{ij}\Phi(z_i,z_j)_{1}ln(A_{ij})-\sum_{ij}\Phi(z_i,z_j)_{2}ln(1-A_{ij})
\end{align}%
where $\Phi(z_i,z_j)_{1}=W_{ij}$ is the output corresponding to same-class semantic similarity and $\Phi(z_i,z_j)_{2}=1-W_{ij}$ is the output corresponding to different-class semantic dissimilarity. Now we substitute equation (\ref{equ7}) into equation (\ref{equ8}), we arrive at the following expression.
\begin{align}
\label{equ9}
L_{unsup} = &\beta\sum_{ij}W_{ij}\|f(x_i)-f(x_j)\|^2\nonumber\\
&-\sum_{ij}(1-W_{ij})ln(1-\exp(-\beta \|f(x_i)-f(x_j)\|^2))
\end{align}%
The first term of equation is actually the traditional graph Laplacian regularization in equation (\ref{equ2}),  with an additional parameter $ \beta $. It penalizes the smoothness of $f$ over the latent graph $G_l$.  The second term encourages dissimilar nodes on the graph by \textit{forcibly pulling} $f(x_i)$ and $f(x_j)$ far apart.  This penalty of the discrepancy between semantic graph $G_s$ and confidence graph $G_c$ results in an extension to the traditional graph Laplacian regularization. By including the second dissimilarity term, the new regularizer naturally prevents from model collapse (i.e. setting all output $W_{ij}$  of similarity network  to 0 to obtain a trivial solution to minimize traditional graph Laplacian regularizer).  Note that in $Augment(x_i)$ and $x_i$ can be treated as a pair of similar nodes in semantic graph, so we make $ W \equiv 1 $ for this case. 

The ideal case for categorical labels $ f(\cdot) $ and similarity $ W $ is $ W=1 $ for samples in the same class and $ W=0 $ for samples in different classes. It's not difficult to find that equation (\ref{equ9}) can attain the optimal value in this case. Specifically, when $ W_{ij} $ is  large (small) close or equal to 1 (0) , then the Euclidean distance $\|f(x_i)-f(x_j)\|$ is encouraged to be smaller (larger), and vice versa. We argue that categorical label learning and the semantic similarity learning with the extended graph Laplacian regularizer can promote each other \textit{bidirectionally} by minimizing equation (\ref{equ9}). As demonstrated in experiments, samples from each class are encouraged to form compact, well separated clusters. Since there are sample pairs of individual sample with its perturbed version and different samples in our mini-batch (see equation (\ref{equ11})), both local consistency and global consistency are guaranteed.

\subsubsection{Perturbation Regularization of Similarity}

We also encourage the consistency between a sample pair $ \Phi(z_i, z_j) $ and its perturbed version $ \Phi(z_i', z_j') $ to learn more accurate pairwise similarity, where $ (z_i,z_j) = (g(x_i), g(x_j)) $ and $ (z_i',z_j') = (g(x_i'), g(x_j')) $. The consistency loss of similarity is formulated in equation (\ref{equ10}).
\begin{align}
\label{equ10}
L_{cons} =& \sum_{ij}\|\Phi(z_i, z_j)-\Phi(z_i', z_j')\|^2\nonumber\\
=& \sum_{ij}\|\Phi_{\alpha}(g(x_i), g(x_j))-\Phi_{\alpha'}(g(x_i'), g(x_j'))\|^2
\end{align}%
$ \alpha' $ here is an exponential moving average of network parameters similar to \cite{tarvainen2017mean} and we do not propagate gradients when computing $ \Phi_{\alpha'}(\cdot) $. Since similarity network learns a relationship of a combination of two samples, the learning space is much larger than categorical learning. This perturbation regularization of similarity learning is important to narrow the searching space. We will demonstrate the necessity of adding $ L_{cons} $ in our ablation study.

\subsection{Training Configurations}

Given the above parts, we describe a training strategy to integrate them into a unified semi-supervised learning framework. Since we are training the model for an end-to-end learning of semantic similarity, we organize training batches in the form of \textit{sample pairs} $(x_i, x_j)$ (which means a batch of size $b$ means there are $b$ pairs $(x_i, x_j)$ ). For one batch update, the overall objective function is given in equation (\ref{equ12}).
\begin{align}
	\label{equ12}
	L&= \frac{1}{|B_1|+|B_2|\times 2}(L_{sup\_f}+L_{sup\_W})\nonumber\\
	&+\frac{\lambda_{1}}{|B_1|+|B_2|}L_{unsup}+\frac{\lambda_{2}}{|B_3|}L_{unsup}\nonumber\\
	&+\frac{\lambda_{3}}{|B_1|+|B_2|  + |B_3|}L_{cons}
\end{align}%
where $\lambda_{1}, \lambda_2$ and $\lambda_{3}$ are regularization coefficients. $B_1, B_2$ and $B_3$ are child batch size.  To compute the losses, we divide each training batch into three small child batches of size $B_1, B_2$ and $B_3$, respectively, and they are constructed as follows. We first randomly select $B_1$ samples $ \mathcal{X}_{1} $ from the whole dataset, together with their augmented version $ \mathcal{X}'_1 $, to make the first small batch of size $B_1 $. Then, we randomly select two subsets $ \mathcal{X}_{l1} $ and $ \mathcal{X}_{l2} $ from $ \mathcal{X}_L $ (together with their labels $ \mathcal{Y}_{l1} $  and $ \mathcal{Y}_{l2} $ from $ \mathcal{Y}_{L} $), each of size $B_2$, to make the second small batch. For the third one, we randomly divide $\mathcal{X}_1$ into two equal subsets  $\mathcal{X}_2$ and $\mathcal{X}_3$ to make the third small batch of size $ B_1 / 2 $. We define the structure of one training batch as
\begin{align}
\label{equ11}
batch&:=\nonumber\\
\{&batch_1: (\mathcal{X}_1,\mathcal{X}'_1);\nonumber\\
&batch_2: (\mathcal{X}_{l1},\mathcal{X}_{l2});\nonumber\\
&batch_3: (\mathcal{X}_2,\mathcal{X}_3)\}
\end{align}%
Obviously, $ W \equiv 1 $ for all pairs in $ batch_1 $, so it can be used to evaluate $ L_{sup\_W} $, $ L_{cons} $ and $ L_{unsup} $ (with $W=1$). For $ batch_2 $, we could evaluate all the loss terms, including $ L_{sup\_f},L_{sup\_W}, L_{unsup}$ and $ L_{cons} $. For $ batch_3 $, we only evaluate the unsupervised term $ L_{unsup}$ and $L_{cons} $. Note that $ W $ in $ L_{unsup} $ in $ batch_3 $ is network's output.  

With this training strategy, the classification network and the similarity learning network are optimized simultaneously using both labeled data  and unlabeled data and the similarity is learned in an end-to-end semi-supervised way. In summarize, the full algorithm is shown in Algorithm~\ref{alg1}.

\begin{algorithm}[tb]
	\caption{End-to-End Semi-Supervised Similarity Learning}
	\label{alg1}
	\hspace*{-0.3cm} {\bf Input:}
	$ (x_i, x_j) $ := training batches in equation (\ref{equ11}); $ (\mathcal{Y}_{l1}, \mathcal{Y}_{l2}) $ := one-hot labels of $ (\mathcal{X}_{l1},\mathcal{X}_{l2}) $ in $ batch_2 $; $ W_1 \equiv 1 $ for $ batch_1 $; $ W_2 $ := corresponding one-hot similarity labels of $ batch_2 $; $ f_{\theta}(x), \Phi_{\alpha}(g(x)) $ := neural network with parameters $ \theta, \alpha $.\\
	\hspace*{-3cm} {\bf Parameter:}
	Coefficients $ \beta, \lambda_{1}, \lambda_{2}, \lambda_{3} $.\\
	\begin{algorithmic}[1] 
		\FOR {epoch in $ [1,...,T] $}
		\FOR {mini-batch in $ [1,...,B] $}
		\STATE Calculate each batch's corresponding loss in equation (\ref{equ12})\\
		\STATE update $ \theta,\alpha $ using Adam optimizer~\cite{kingma2015adam}
		\ENDFOR
		\ENDFOR
		\STATE \textbf{return} trained parameters $ \theta, \alpha $
	\end{algorithmic}
\end{algorithm}


\section{Experiments}

In this section, we evaluate the effectiveness of our proposed method on several standard benchmarks and compare the results with the recently reported ones in the literature to show its performance superiority\footnote{The source code to our reproduce experimental results is available at google drive: https://drive.google.com/open?id=1BU-w3pSeIyP4X2--wFM5xO8HpgojxCBN}.

\subsection{A Toy Dataset Experiment}
As an illustrative example, we first evaluate our model on the ``two moons'' and ``two circles'' toy datasets. Each datasets contain 6000 samples of $ x \in \mathcal{R}^2, y \in \{1,2\} $ with Gaussian noise of $\sigma =0.15 $ and $\sigma =0.3 $, respectively. There are 12 labeled samples in ``two moons'' and 8 labeled samples in ``two circles''.  We use a three-layer fully-connected network with a hidden layer of 100 neurons followed by leaky RELU $ \alpha = 0.1 $. Then we concatenate two 100-dimensional vectors to form a new 200-dimensional vector as $ \Phi(\cdot) $'s input.  We  define $ \Phi(\cdot) $ as a three-layer fully-connected network: $ 200 \rightarrow 512 \rightarrow dropout(0.2) \rightarrow 128 \rightarrow dropout(0.2) \rightarrow 64 \rightarrow 2 $.

\begin{figure}[ht]
	\centering
	\subfigure[MT]{\includegraphics[width=0.2\textwidth]{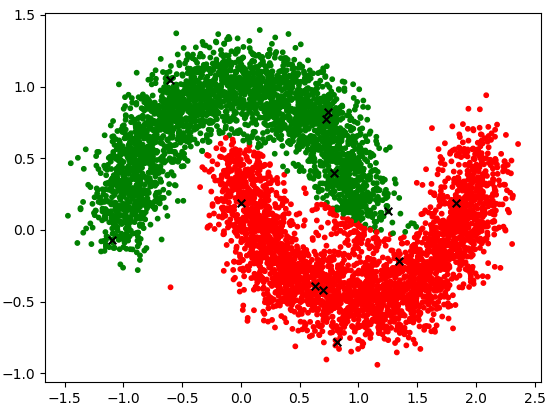}}
	\subfigure[$\Pi$]{\includegraphics[width=0.2\textwidth]{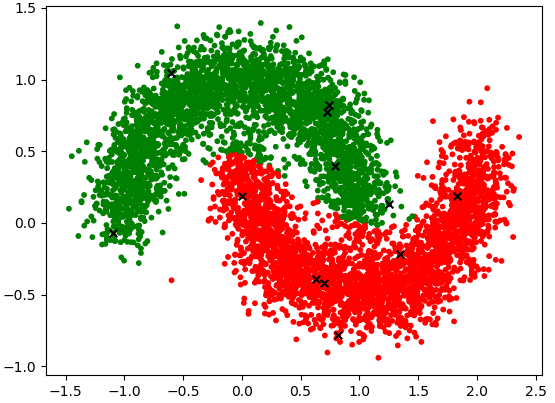}}
 	\subfigure[SNTG]{\includegraphics[width=0.2\textwidth]{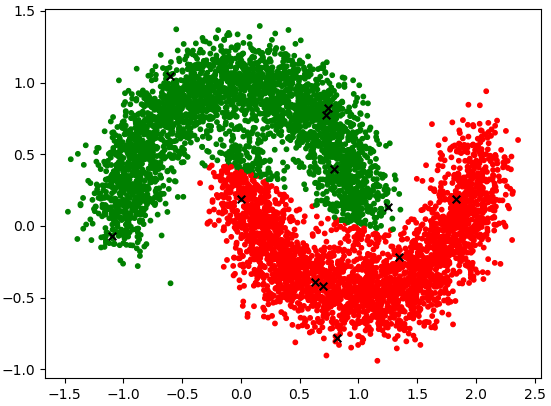}}
 	\subfigure[Ours]{\includegraphics[width=0.2\textwidth]{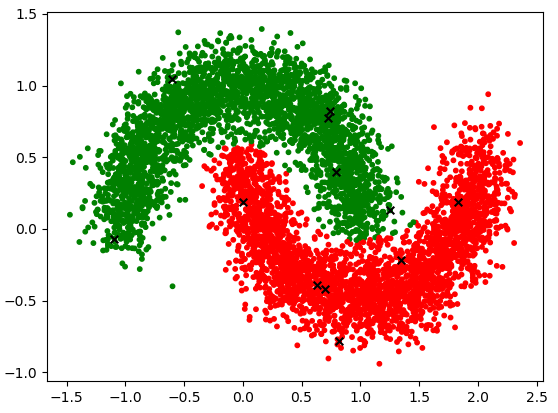}}
	\caption{Classification results of our method and baseline methods on the ``two moons'' dataset. 12 labeled samples are marked with the black cross. Note the inside end of each moon.}
	\label{fig2}
\end{figure}

\begin{figure}[ht]
	\centering
	\subfigure[MT]{\includegraphics[width=0.2\textwidth]{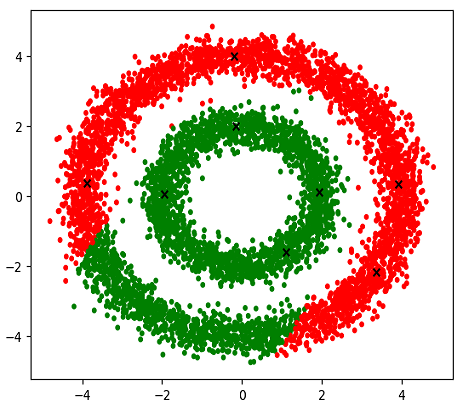}}
	\subfigure[$\Pi$]{\includegraphics[width=0.2\textwidth]{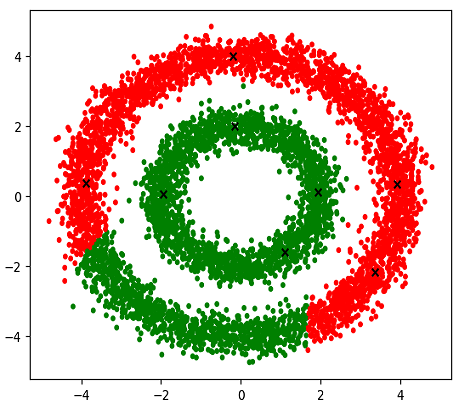}}
	\subfigure[SNTG]{\includegraphics[width=0.2\textwidth]{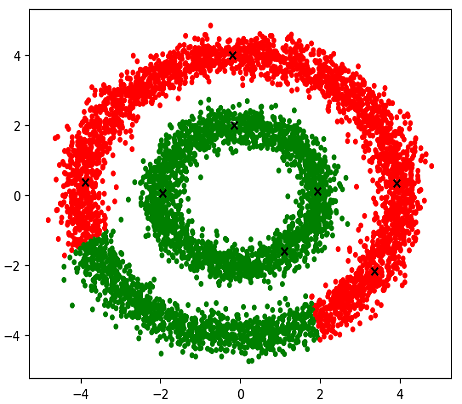}}
	\subfigure[Ours]{\includegraphics[width=0.2\textwidth]{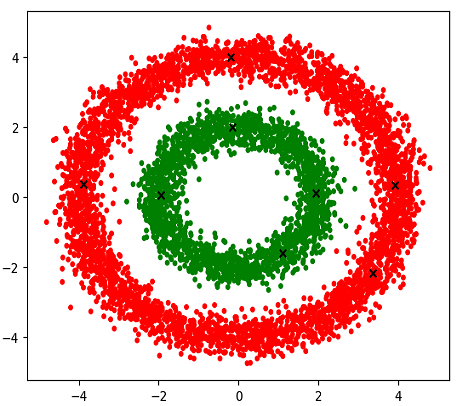}}
	\caption{Classification results of our method and baseline methods on the ``two circles'' dataset. 8 labeled samples are marked with the black cross.}
	\label{fig22}
\end{figure}

We compare our method with MT \cite{tarvainen2017mean}, $\Pi$ \cite{laine2017temporal} and STNG \cite{luo2018smooth}. The results are depicted in Figure~\ref{fig2} and \ref{fig22}.  From the figure we can see that due to the irregular distribution of the labeled samples and the considerable class mixture,  the baseline algorithms have a considerable amount of  misclassified points at the inside end of each moon. In contrast, our method corrects the prediction of these samples and has a better performance.

\subsection{Benchmark Datasets Experiments}
We evaluate the classification performance of the proposed model and compare its results with several recently developed SSL models in \cite{laine2017temporal,tarvainen2017mean,miyato2018virtual,luo2018smooth,verma2019interpolation,berthelot2019mixmatch,berthelot2019remixmatch,taherkhani2019matrix,iscen2019label,athiwaratkun2019there}. 
In each experiment with different number of labels on different datasets, we run our model for 5 times across different random data splits and report the mean and standard deviation of the test error rate. Results of baseline algorithms are adopted directly from the original papers if they are available, or run by us using the provided code and suggested parameters if not available.

\subsubsection{Datasets and Preprocessing}

We conduct experiments on three datasets widely-used in previous SSL studies: SVHN, CIFAR-10 and CIFAR-100. We randomly choose a small part of the training samples as labeled and use the rest training data as unlabeled. Following the common practice in the literature such as \cite{laine2017temporal,tarvainen2017mean,luo2018smooth}, we ensure that the number of labeled samples between all classes are balanced and perform standard augmentation (including 2-pixel random translation all datasets and random horizontal flip on CIFAR-10/100).

\paragraph{SVHN}

The SVHN dataset includes 73257 training samples and 26032 test samples of size $ 32 \times 32 $. The task is to recognize the centered digit (0-9) of each image. For SVHN, we use the same standard augmentation and pre-processing as those in prior work~\cite{laine2017temporal,tarvainen2017mean}.

\paragraph{CIFAR-10.}

The CIFAR-10 dataset consists of 60000 RGB images of size $ 32 \times 32 $ from 10 classes. There are 50000 training samples and 10000 test samples. For CIFAR-10, we first normalize the images using per-channel standardization. Then we augment the dataset by random horizontal flips with probability 0.5 and random translation with 2 pixels. Unlike prior work~\cite{laine2017temporal}, we found it is not necessary to use ZCA whitening, nor add Gaussian noise to the input images.

\paragraph{CIFAR-100.}

The CIFAR-100 dataset is just like the CIFAR-10, except that it has 100 classes containing 600 images per class. There are also 50000 training samples and 10000 test samples. We use the same data normalization method as CIFAR-10. But we evaluate the performance on CIFAR-100 only with RandAugment since it is a more difficult classification task.

Furthermore, we also perform experiments with RandAugment on SVHN and CIFAR-10 to achieve better results, which will be shown in section~\ref{results}.

\subsubsection{Implementation Details}

\paragraph{Settings.}
For categorical labels learning (from the input images $ \mathcal{X} $ to classification output $ f(\cdot) $), we use the standard ``CNN-13'' architecture that has been employed as a common network structure in recent perturbation-based SSL approaches~\cite{laine2017temporal,tarvainen2017mean,luo2018smooth}. We treat the 128-dimensional vector before the linear classifier as the logits output of the feature extractor $ z = g(\cdot) $. Then we concatenate two features to form a new 256-dimensional vector as the similarity network $ \Phi(\cdot) $'s input. We define $ \Phi(\cdot) $ as a four-layer fully-connected network: $ 256 \rightarrow 512 \rightarrow dropout(a) \rightarrow 512 \rightarrow dropout(a) \rightarrow 128 \rightarrow dropout(a) \rightarrow 64 \rightarrow 2 $. Experiments on all three datasets are performed and results are recorded for comparison with baseline algorithms.

\paragraph{Parameters.} The hyperparameters of our method include $ \beta, \lambda_{1}, \lambda_{2}, \lambda_{3} $. Following SNTG \cite{luo2018smooth}, we also use a ramp-up schedule for both the learning rate and the coefficients $ \lambda_{1}, \lambda_{2}, \lambda_{3} $ in the beginning. Since neither the categorical labels nor the similarity is accurate at the beginning of training, we do not add the third term of equation (\ref{equ12}) until 100 epochs. Moreover, we define one epoch when $ x_1 $ traverses all samples from the dataset. 

In each iteration, we sample a mini-batch according to equation~(\ref{equ11}), where we set $ |B_1|=100, |B_2|=10 $ and $ |B_3|=50 $. Following~\cite{oliver2018realistic}, we select the best hyperparameters for our method using a validation set of 1000, 5000 and 5000 labeled samples for SVHN, CIFAR-10 and CIFAR-100 respectively. For coefficients $ \lambda_{1} $ and $ \lambda_{2} $, we set them as $ k_{1} \times \dfrac{n\_labeled}{n\_training} $ and $ k_{2} \times \dfrac{n\_labeled}{n\_training} $. Then we only need to adjust $ k_{1} $ and $ k_{2} $. For SVHN, we set $ \beta = 1.5$, $k_{1} = 2 k_{2} = 8, \lambda_{3} = 0.05 $, and we run the experiments for 500 epochs. For CIFAR-10 with standard augmentation, we set $ \beta = 3.0, k_{1} = k_{2} = 3, \lambda_{3} = 0.15 $ and $ dropout\_rate = 0.2 $. And we use $ \beta = 3.0, k_{1} = 2 k_{2} = 3 | 6 $ (for different number of labels), $ \lambda_{3} = 0.05 $ and $ dropout\_rate = 0 $ instead for CIFAR-10 with RandAugment. Note that we remove dropout here because we found RandAugment is a more effective perturbation than dropout on CIFAR-10 dataset. For CIFAR-100, we set $ \beta = 3.0, k_{1} = 2k_{2} = 1.5, \lambda_{3} = 0.15 $ and $ dropout\_rate = 0 $. we run the experiments for 600 epochs for both CIFAR-10 and CIFAR-100. The coefficients $ \lambda_{1}, \lambda_{3} $ are ramped up from 0 to their maximum value at first 80 epochs. Besides, $ \lambda_{2} $ is 0 at first 100 epochs, then it is ramped up to its maximum value in the next 50 epochs.

Except for the parameters stated above, all other hyperparameters remain unchanged from MT implementation~\cite{tarvainen2017mean}.

\paragraph{Advanced Data Augmentation}
To explore the performance bound of our method, we also combine our SSL with advanced data augmentation. We use the similar augmentation strategy as reported in RandAugment~\cite{cubuk2019randaugment}, e.g. randomly add two different strong augmentations to each image in $ \mathcal{X}_1 $ with random magnitude. Then $ f(\mathcal{X}_1) $ in $ L_{unsup} $ is forced to learn the fixed target $ f(\mathcal{X}') $, which can be seen as teacher model's output and only standard augmentation is added to $ \mathcal{X}'_1 $.


\subsubsection{Evaluation of Classification Accuracy}\label{results}

For SVHN, we evaluate the error rate with 250, 500 and 1000 labeled samples respectively, and experimental results of standard augmentation and RandAugment are presented in Table~\ref{tab1} and Table~\ref{tab2}.  Notably, while the error rates of MT+STNG decrease 0.06, 0.19, 0.09 (for 250, 500 and 1000 labels, receptively) comparing to that of its baseline MT, the error rate drop of our method is 0.31, 0.31 and 0.38 comparing to MT+STNG.  This suggests that learned similarity is much better than simply assigning 0-1 similarity using pseudo labels. In both tables, it can be seen that our method outperforms the baseline algorithms by a considerable margin on the SVHN dataset.

\begin{table}[htbp!]
	\caption{Error rates (\%) on SVHN with standard augmentation (bottom 6 rows are graph-based SSL; $ ^\ast $ no standard augmentation).}
	\label{tab1}
	\centering
	\begin{tabular}{lrrr}
		\toprule
		Method    & 250 labels & 500 labels & 1000 labels    \\
		\midrule
		$ \Pi $ model~\cite{laine2017temporal}    & 9.93 $ \pm $ 1.15 & 6.65 $ \pm $ 0.53 & 4.82 $ \pm $ 0.17    \\
		TempEns~\cite{laine2017temporal}     & 12.62 $ \pm $ 2.91 & 5.12 $ \pm $ 0.13 & 4.42 $ \pm $ 0.16    \\
		VAT~\cite{miyato2018virtual}    & – & – & 5.42 $ \pm $ 0.22    \\
		MT~\cite{tarvainen2017mean}    & 4.35 $ \pm $ 0.50 & 4.18 $ \pm $ 0.27 & 3.95 $ \pm $ 0.19    \\ \midrule
		CCLP \cite{kamnitsas2018semi}$ ^\ast $ & - & - & 5.69 $ \pm $ 0.28 \\
		GSCNN~\cite{taherkhani2019matrix}    & – & – & 5.13 $ \pm $ 0.39    \\
		LPDSSL~\cite{iscen2019label}    & { 18.45} & 9.49 & 7.38    \\
		$ \Pi $+SNTG~\cite{luo2018smooth}    & 5.07 $ \pm $ 0.25 & 4.52 $ \pm $ 0.30 & 3.82 $ \pm $ 0.25    \\
		MT+SNTG~\cite{luo2018smooth}    & 4.29 $ \pm $ 0.23 & 3.99 $ \pm $ 0.24 & 3.86 $ \pm $ 0.27    \\
		Ours   & \textbf{3.98 $ \pm $ 0.21} & \textbf{3.68 $ \pm $ 0.16} & \textbf{3.48 $ \pm $ 0.11}    \\
		\bottomrule
	\end{tabular}
\end{table}

\begin{table}[htbp!]
	\caption{Error rates (\%) on SVHN with RandAugment \cite{cubuk2019randaugment}. $ \dagger $ denotes a different architecture WRN-28-2 applied.}
	\label{tab2}
	\centering
	\resizebox{245pt}{42pt}
	{\begin{tabular}{lrrr}
		\toprule
		Method    & 250 labels & 500 labels & 1000 labels    \\
		\midrule
		ICT~\cite{verma2019interpolation}    & 4.78 $ \pm $ 0.68 & 4.23 $ \pm $ 0.15 & 3.89 $ \pm $ 0.04    \\
		MixMatch~\cite{berthelot2019mixmatch}$ \dagger $    & 3.78 $ \pm $ 0.26 & 3.64 $ \pm $ 0.46 & 3.27 $ \pm $ 0.31    \\
		ReMixMatch~\cite{berthelot2019remixmatch}$ \dagger $    & 3.10 $ \pm $ 0.50 & – & 2.83 $ \pm $ 0.30    \\
			MT+SNTG~\cite{luo2018smooth}    & 2.60 & 2.57 & 2.55    \\
		\midrule
		Ours    & \textbf{2.42 $ \pm $ 0.33} & \textbf{2.31 $ \pm $ 0.10} & \textbf{2.26 $ \pm $ 0.04}    \\
		\bottomrule
	\end{tabular}}
\end{table}

For CIFAR-10 with standard augmentation, we report the results with 1000, 2000 and 4000 labels. We also conduct experiments for fewer labeled samples from 250 to 4000  with a stronger data augmentation \cite{cubuk2019randaugment}, since models are more likely to overfit fewer labeled samples on CIFAR-10. Results on CIFAR-10 are listed in Table~\ref{tab3} and Table~\ref{tab4}. From these tables we can see that our method  outperforms prior works for most cases. As an interesting trend in table \ref{tab4} for strong augmentation, the performance gains of our model grows larger as the number of labeled samples becomes smaller. Compared to prior works, the error rate of our model decreases an amount of 2.59\% for 250 labels.

\begin{table}[htbp!]
	\caption{Error rates (\%) on CIFAR-10 with standard augmentation  (bottom 6 rows are graph-based SSL; $ ^\ast $ no standard augmentation).}
	\label{tab3}
	\centering
	\resizebox{245pt}{57pt}
	{\begin{tabular}{lrrr}
			\toprule
			Method    & 1000 labels & 2000 labels & 4000 labels    \\
			\midrule
			$ \Pi $ model~\cite{laine2017temporal}    & 31.65 $ \pm $ 1.20 & 17.57 $ \pm $ 0.44 & 12.36 $ \pm $ 0.31    \\
			TempEns~\cite{laine2017temporal}    & 23.31 $ \pm $ 1.01 & 15.64 $ \pm $ 0.39 & 12.16 $ \pm $ 0.24    \\
			MT~\cite{tarvainen2017mean}    & 21.55 $ \pm $ 1.48 & 15.73 $ \pm $ 0.31 & 12.31 $ \pm $ 0.28    \\
			VAT~\cite{miyato2018virtual}    & – & – & 11.36 $ \pm $ 0.34    \\
			MixMatch (without Mixup)~\cite{berthelot2019mixmatch}    & 20.16 & 14.32 & 10.97    \\ \midrule
			CCLP \cite{kamnitsas2018semi}$ ^\ast $ & - & - & 18.57 $ \pm $ 0.41 \\
			GSCNN~\cite{taherkhani2019matrix}    & 18.98 $ \pm $ 0.82 & 16.82 $ \pm $ 0.47 & 15.49 $ \pm $ 0.64    \\
			LPDSSL~\cite{iscen2019label}    & 22.02 $ \pm $ 0.88 & 15.66 $ \pm $ 0.35 & 12.69 $ \pm $ 0.29    \\
			$ \Pi $+SNTG~\cite{luo2018smooth}    & 21.23 $ \pm $ 1.27 & 14.65 $ \pm $ 0.31 & 11.00 $ \pm $ 0.13    \\
			TempEns+SNTG~\cite{luo2018smooth}    & \textbf{18.41 $ \pm $ 0.52} & 13.64 $ \pm $ 0.32 & 10.93 $ \pm $ 0.14    \\
			Ours    & 18.47 $ \pm $ 0.57 & \textbf{13.62 $ \pm $ 0.33} & \textbf{10.78 $ \pm $ 0.24}    \\
			\bottomrule
	\end{tabular}}
\end{table}

\begin{table*}[htbp!]
	\caption{Error rates (\%) on CIFAR-10 with RandAugment \cite{cubuk2019randaugment}. $ \dagger $ denotes a different architecture WRN-28-2 applied.}
	\label{tab4}
	\centering
	\begin{tabular}{lrrrrr}
		\toprule
		Method    & 250 labels & 500 labels & 1000 labels & 2000 labels & 4000 labels    \\
		\midrule
		ICT~\cite{verma2019interpolation}    & – & – & 15.48 $ \pm $ 0.78 & 9.26 $ \pm $ 0.09 & 7.29 $ \pm $ 0.02    \\
		MixMatch~\cite{berthelot2019mixmatch}$ \dagger $    & 11.08 $ \pm $ 0.87 & 9.65 $ \pm $ 0.94 & 7.75 $ \pm $ 0.32 & 7.03 $ \pm $ 0.15 & \textbf{6.24 $ \pm $ 0.06}    \\
			MT+SNTG~\cite{luo2018smooth}    & 10.56 & 9.39 & 8.57 & 7.47 & 6.63    \\
		\midrule
		Ours    & \textbf{7.97 $ \pm $ 0.37} & \textbf{7.79 $ \pm $ 0.22} & \textbf{7.32 $ \pm $ 0.17} & \textbf{6.97 $ \pm $ 0.14} & 6.24 $ \pm $ 0.12    \\
		\bottomrule
	\end{tabular}
\end{table*}

For CIFAR-100, we perform experiments with 10000 labels with RandAugment. The results are in Table~\ref{tab5}.  All results use the same CNN-13 architecture. Results on CIFAR-100 again confirm the effectiveness of our method.

\begin{table}[htbp!]
	\caption{Error rates (\%) on CIFAR-100.}
	\label{tab5}
	\centering
	\begin{tabular}{lr}
		\toprule
		Method    & 10000 labels    \\
		\midrule
		$ \Pi $ model~\cite{laine2017temporal}    & 39.19 $ \pm $ 0.36    \\
		TempEns~\cite{laine2017temporal}    & 38.65 $ \pm $ 0.51    \\
		LPDSSL+MT~\cite{iscen2019label}    & 35.92 $ \pm $ 0.47    \\
		MT+fast SWA~\cite{athiwaratkun2019there}    & 33.62 $ \pm $ 0.54\\ \midrule
		Ours    & \textbf{32.25 $ \pm $ 0.44}    \\
		\bottomrule
	\end{tabular}
\end{table}

We also find that the classification performance of our method is relatively stable to the number of labels when strong data augmentation is applied. It indicates that our method with RandAugment could make better use of unlabeled samples when labels are scarce.

\subsubsection{Evaluation of Learned Similarity}
\begin{figure}[htbp!]
	\centering
	\includegraphics[width=0.48\textwidth]{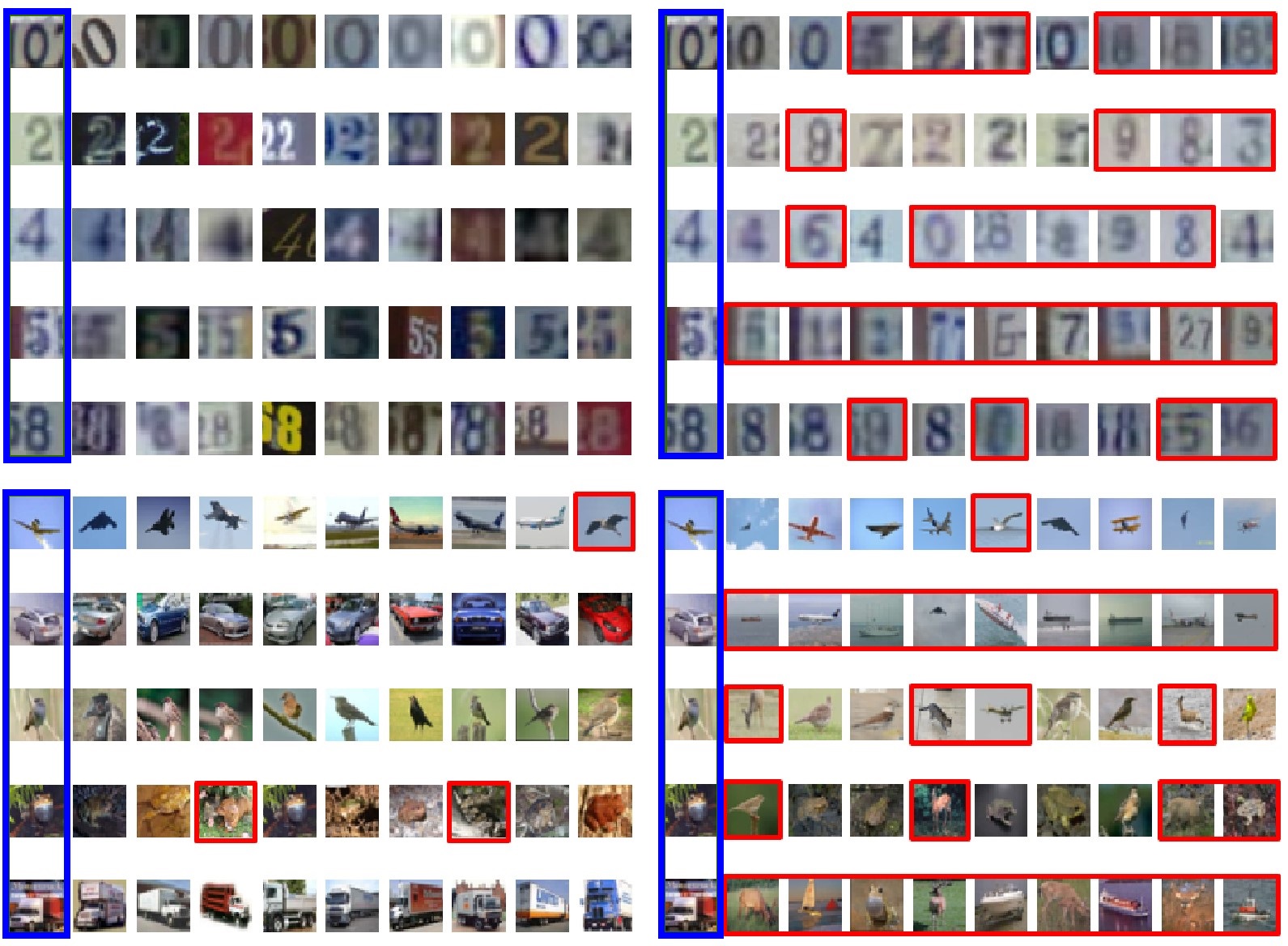}
	\caption{Query results given by our method (left two panels) and the Gaussian kernel function (right two panels) in  SVHN and CIFAR-10. 10 target samples are indicated by the blue rectangle on the left of each panel. Samples surrounded by red rectangles indicate that they come from a different class to the corresponding query target.}
	\label{fig3}
\end{figure}

One extra merit that also differentiates our semi-supervised approach from existing approaches (e.g. \cite{laine2017temporal,tarvainen2017mean,luo2018smooth}) is the similarity evaluation byproduct, which can evaluate directly the semantic-level similarity of any two input samples and thus could be utilized for image comparison/query. After training, the two component networks (feature network and similarity network in Figure \ref{fig1}) could be used as encoding network and similarity evaluation network, respectively.  In this section, we conduct experiments to show that our method can learn high-level semantic similarity information from raw images. We randomly select 5 samples from the testing set of SVHN and CIFAR-10 respectively and show their $k$-nearest neighbors ($k$=9) queried according to the learned similarity and the Gaussian kernel function. Figure~\ref{fig3} displays the nearest neighbors  in descending order.

 The results of our method are obtained using  1000 labeled samples for SVHN and  4000 labeled samples for CIFAR-10 with standard data augmentation. As can be observed, our learned similarity focuses more on the high-level semantic features of the sample contents, ignoring other distraction details and transformations such as rotation, translation, brightness, etc. In addition, our method can measure the relative value of similarity rather than simply assigning 0 or 1 to a pair of samples, as have been done by SNTG \cite{luo2018smooth} and other pseudo label based algorithms.

\begin{figure}[htbp!]
	\centering
	\includegraphics[width=0.48\textwidth]{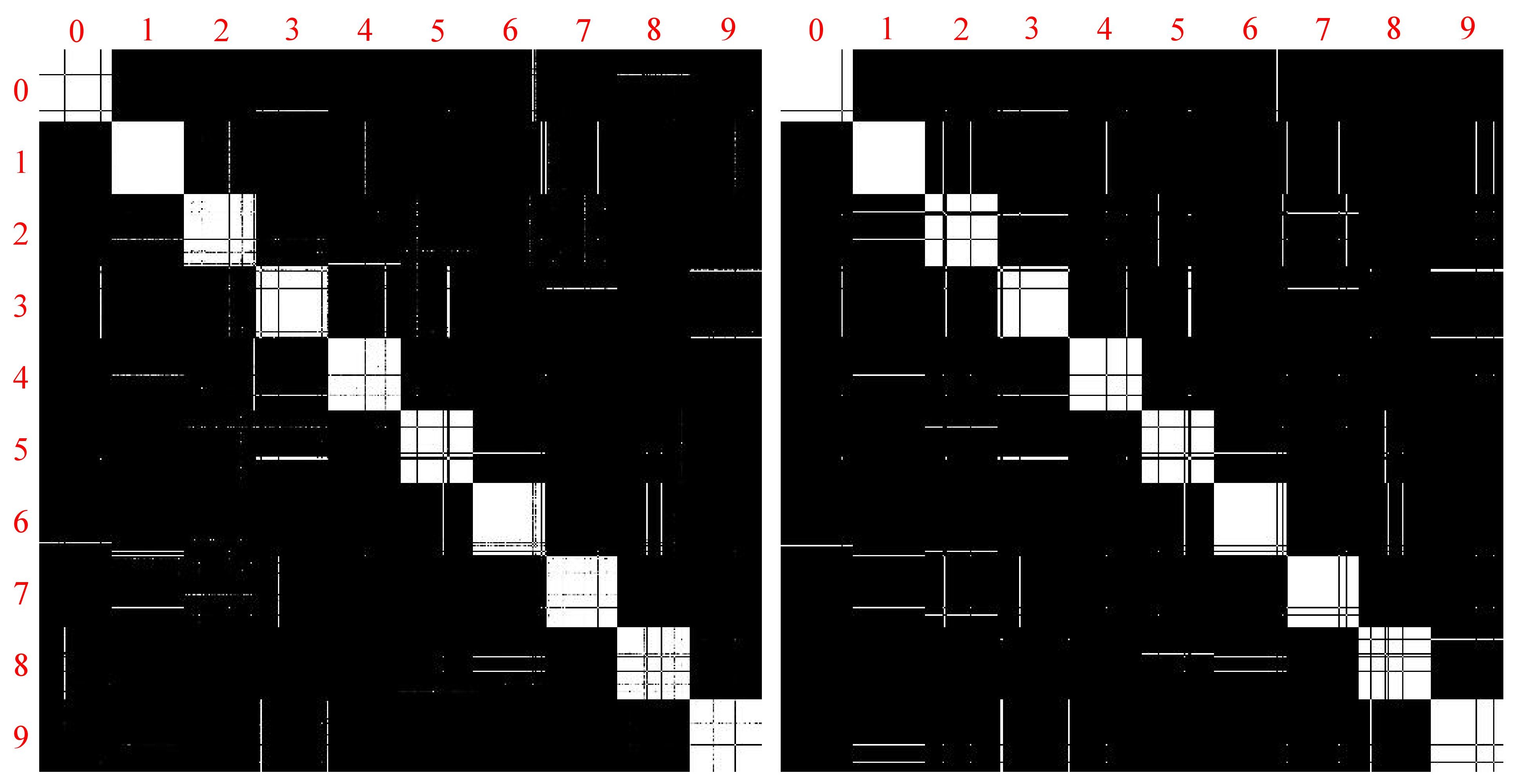}
	\caption{Visualization of similarity matrices. We get the results by running our model (left) and reproducing the $ \Pi $+SNTG model \cite{luo2018smooth} (right) on SVHN with 1000 labels.}
	\label{fig4}
\end{figure}

To quantitatively evaluate the difference between  learned similarity and pseudo-label assignment, we randomly select 500 samples from the testing set of SVHN and plot the similarity matrices given by our method and $ \Pi $+SNTG \cite{luo2018smooth} and depict them in Figure~\ref{fig4}. Note that the elements of the similarity matrix of $ \Pi $+SNTG are either 0 or 1 but ours are in $[0,1]$, having more useful closeness ranking meaning. Visually comparing the two matrices, our model learns more accurate similarity for classes \{2, 3, 4, 7, 8, 9\} than $ \Pi $+SNTG, but worse for classes \{0, 6\}.
Quantitatively, the mean squared error (MSE) between the two matrices and the ideal similarity matrix (block diagonal) are 0.149 (ours) and 0.201 ($ \Pi $+SNTG), respectively.

\subsection{Ablation Study}

Finally, we conduct experiments to investigate the effectiveness of the similarity network  on semi-supervised learning. We employ the $\Pi$ \cite{laine2017temporal} and MT \cite{tarvainen2017mean} as the base model to perform the ablation study on SVHN with 1000 labels. The results are shown in Table~\ref{tab6}. From these results, one can see that the similarity learning is beneficial to improve classification performance for the models.

\begin{table}[htbp!]
	\caption{Ablation study. Error rates (\%) are reported on SVHN with 1000 labels.}
	\label{tab6}
	\centering
	\begin{tabular}{lr}
		\toprule
		Ablation    & 1000 labels    \\
		\midrule
		\textit{without} learning similarity ($ \Pi $ model)    & 4.82    \\
		\textit{with} learning similarity ($ \Pi $ model)    & 3.84    \\ \hline	
		\textit{without} learning similarity (MT)    & 3.93    \\
		\textit{with} learning similarity (MT)    & 3.50    \\
		\bottomrule
	\end{tabular}
\end{table}

\section{Conclusions and Discussions}

In this paper, we proposed an end-to-end semi-supervised similarity learning approach to jointly optimize a categorical labeling network and a similarity measure network to minimize an overall semi-supervised objective function. Experiments on three widely used image benchmark datasets show that our method outperforms or is comparable to other graph-based SSL methods and can learn more accurate similarity. With advanced data augmentation, our method is able to fully exploit the data information to achieve state-of-the-art results. After training, an extra reward of our model is the similarity network, which could be used potentially for semantic-level image query. It is also worth mention that our method is extendable and is easy to apply to other methods by adding a neural network to learn the similarity. Thus in future work, we will further exploit the capability of our method on other learning tasks, such as image retrieval.

A potential limitation of our similarity learning approach is that the learned similarity matrix cannot be guaranteed to be positive semidefinite (PSD), which may restrict its application on some learning tasks, e.g. kernel based learning algorithms. Nevertheless, indefinite kernel learning has been found to be interesting \cite{ying2009analysis,loosli2015learning} and we may try to combine our approach with the indefinite kernel learning algorithms in the future.

%

\bibliographystyle{ACM-Reference-Format}
\bibliography{sample-base}



\end{document}